%% file: main.tex
\documentclass[dvipsnames]{article} %
\usepackage{colm2024_conference}

\usepackage{booktabs}
\usepackage{graphicx}
\usepackage{enumitem}
\usepackage{wrapfig}
\usepackage{algorithm}
\usepackage{algpseudocode}

\usepackage{microtype}
\usepackage{amsmath}
\usepackage{colortbl}
\usepackage[utf8]{inputenc}
\definecolor{lightgray}{rgb}{0.9,0.9,0.9}
\usepackage{caption}
\usepackage{subcaption}
\usepackage{setspace}
\usepackage{url}
\usepackage{multirow}
\usepackage{colortbl}
\usepackage{tabularx}
\usepackage{blindtext}
\usepackage{pgfplots}
\pgfplotsset{compat=1.18} 
\usepackage{tikz}
\usetikzlibrary{er,positioning,bayesnet}
\usepackage{makecell}
\usepackage{tipa}
\usepackage{siunitx}
\usepackage{nicefrac}
\usepackage{tocloft}
\usepackage{listings}
\usepackage[raster,skins]{tcolorbox} %
\usepackage{xltabular}
\usepackage{adjustbox}
\usepackage{xurl}
\usepackage{rotating}
\usepackage[normalem]{ulem}
\usepackage{kotex} 
\useunder{\uline}{\ul}{}

\input{math_commands.tex}

\newcommand*\justify{%
  \fontdimen2\font=0.4em
  \fontdimen3\font=0.2em
  \fontdimen4\font=0.1em
  \fontdimen7\font=0.1em
  \hyphenchar\font=`\-
}

\renewcommand{\texttt}[1]{%
  \begingroup
  \ttfamily
  \begingroup\lccode`~=`/\lowercase{\endgroup\def~}{/\discretionary{}{}{}}%
  \begingroup\lccode`~=`[\lowercase{\endgroup\def~}{[\discretionary{}{}{}}%
  \begingroup\lccode`~=`.\lowercase{\endgroup\def~}{.\discretionary{}{}{}}%
  \catcode`/=\active\catcode`[=\active\catcode`.=\active
  \justify\scantokens{#1\noexpand}%
  \endgroup
}

\title{A.X K1 Technical Report}

\author{
\bf SK Telecom
}

\begin{document}

\maketitle

\input{abstract}

\input{introduction}

\input{architecture}

\input{pretraining}

\input{posttraining}

\input{evaluation}

\input{conclusion}

\appendix
\input{authors}
\input{appendix}

\clearpage
\bibliography{references}
\bibliographystyle{colm2024_conference}

\end{document}

%% file: math_commands.tex
\usepackage{amsmath,amsfonts,bm}

\def\eqref#1{equation~\ref{#1}}

\def\1{\bm{1}}

\DeclareMathAlphabet{\mathsfit}{\encodingdefault}{\sfdefault}{m}{sl}
\SetMathAlphabet{\mathsfit}{bold}{\encodingdefault}{\sfdefault}{bx}{n}



%% file: abstract.tex
\begin{abstract}
We introduce A.X K1, a 519B-parameter Mixture-of-Experts (MoE) language model trained from scratch. Our design leverages scaling laws to optimize training configurations and vocabulary size under fixed computational budgets. A.X K1 is pre-trained on a corpus of approximately 10T tokens, curated by a multi-stage data processing pipeline. Designed to bridge the gap between reasoning capability and inference efficiency, A.X K1 supports explicitly controllable reasoning to facilitate scalable deployment across diverse real-world scenarios. We propose a simple yet effective Think-Fusion training recipe, enabling user-controlled switching between thinking and non-thinking modes within a single unified model. Extensive evaluations demonstrate that A.X K1 achieves performance competitive with leading open-source models, while establishing a distinctive advantage in Korean-language benchmarks.
\end{abstract}

%% file: introduction.tex
\section{Introduction}

Large language models (LLMs) have significantly progressed in real-world applications such as chatbots~\citep{achiam2023gpt, comanici2025gemini}, search engines~\citep{perplexity_ai, xiong2024search_llm}, and coding assistants~\citep{guo2024deepseek, hui2024qwen2}. In parallel, LLMs have rapidly advanced in reasoning capability through scaling model parameters and reinforcement learning~\citep{kimiteam2025kimik2openagentic, deepseekai2025deepseekr1, yang2025qwen3technicalreport}. These advances are largely driven by increased model capacity~\citep{comanici2025gemini, deepseekai2025deepseekr1} and expanded test-time computation~\citep{snell2024scaling, muennighoff2025s1}. However, translating such capabilities into practical deployments introduces several fundamental challenges.

First, achieving strong reasoning performance under practical inference constraints requires carefully balancing massive knowledge capacity with deployment feasibility. To address this challenge, we design the architecture of our large-scale language model by referencing recent scaling laws for Mixture-of-Experts (MoE) models~\citep{tian2025greaterleveragescalinglaws} and vocabulary size~\citep{tao2024scaling}. In particular, the model is configured as a 519B-parameter MoE architecture with 33B active parameters, pushing model capacity within practical hardware constraints, and adopts an optimized vocabulary size of 160K. This architectural choice prioritizes inference throughput and knowledge density over strictly compute-optimal training efficiency, reflecting the requirements of real-world, large-scale deployment.

Second, as LLMs are increasingly deployed in real-world systems, an efficiency challenge arises: the reasoning process relies on extensive token generation, causing prohibitive latency and inference costs due to the thinking process~\citep{yang2025qwen3technicalreport, deepseekai2025deepseekr1}. While the thinking process demonstrates superior performance on complex tasks, applying deep reasoning to simple queries results in unnecessary computational waste and degraded user experience due to slow response times. Thus, practical applications necessitate \textit{flexible control over the extent of reasoning}, allowing systems to balance performance with computational efficiency rather than indiscriminately maximizing reasoning depth. To address this, we introduce a simple yet effective \textit{Think-Fusion} training recipe that enables A.X K1 to support both explicit thinking and standard (direct) modes within a single unified model. This capability empowers users to dynamically adjust the computational cost based on task complexity, bridging the gap between high-capability reasoning and scalable real-world deployment.

In this work, we present \textbf{A.X K1}, \textit{a compute-efficient Mixture-of-Experts (MoE) language model trained from scratch} to address the above challenges. Grounded in MoE and vocabulary scaling laws~\citep{tian2025greaterleveragescalinglaws, tao2024scaling}, A.X K1 is configured with 519B total parameters and 33B active parameters, enabling high-capacity modeling under a fixed computational budget. Beyond architectural scale, A.X K1 explicitly supports both reasoning-intensive and direct inference modes through a unified training framework, allowing users to control computational cost at inference time. As a result, the model achieves competitive performance among leading open-weight peers with comparable active parameter sizes across a broad range of benchmarks, including knowledge, instruction following, mathematics, and coding.

Beyond its technical contributions, A.X K1 represents a strategic effort to establish a robust \textit{Sovereign AI} ecosystem. By developing a high-performance foundation model that deeply understands the linguistic and cultural characteristics of Korea, we aim to reduce reliance on foreign proprietary models and accelerate the adoption of trustworthy AI across local industries, government, and academia. This initiative aligns with our broader objective of democratizing advanced reasoning capabilities and tailoring them to the specific needs of the Korean AI ecosystem.

We highlight three key aspects of A.X K1 as follows:
\begin{enumerate}[label=(\arabic*), leftmargin=*]
\item \textbf{Compute-Efficient Pre-training:} We demonstrate large-scale MoE pre-training with 519B total parameters using approximately 10T tokens, guided by scaling laws. Despite a fixed compute budget, the model approaches the performance of larger-scale systems trained with substantially more data, indicating strong computational efficiency and further headroom with additional resources.
\item \textbf{Flexible Reasoning via Think-Fusion:} We introduce a methodology for constructing a practical unified model employing our \textit{Think-Fusion} recipe. By synergizing \textit{Model Merging} and \textit{Think-Fusion SFT}, we efficiently unify hybrid capabilities within a single architecture. The resulting model supports a \textit{non-thinking mode} for concise, low-latency responses and a \textit{thinking mode} for extensive reasoning chains, empowering users to flexibly optimize resource allocation based on task complexity.
\item \textbf{Frontier-Scale Engineering Capability:} A.X K1 demonstrates the feasibility of end-to-end engineering for 500B+ parameter models, including stable training, reproducible pipelines, and system-level optimizations. These efforts establish a foundation for scaling toward frontier-class model development.
\end{enumerate}

Extensive evaluations show that A.X K1 matches or exceeds state-of-the-art open-source baselines across English and Korean benchmarks, while providing explicit, user-controllable switching between thinking and non-thinking modes at inference time. Beyond benchmarks, our results validate the feasibility of unifying implicit and explicit reasoning processes within a single parameter space. We release A.X K1\footnote{https://huggingface.co/skt/A.X-K1}, positioning it as a practical, controllable foundation model.

\paragraph{Organization.} The remainder of this report details the model architecture and tokenizer design (Sec.~\ref{sec:architecture}), pre-training (Sec.~\ref{sec:pretrain}), post-training (Sec.~\ref{sec:posttrain}), and evaluation results (Sec.~\ref{sec:evaluation}).

%% file: architecture.tex
\section{Architecture}
\label{sec:architecture}

\begin{table}[htbp]
\caption{Architecture Overview of A.X K1 MoE Model,  which adopts Multi-head Latent Attention (MLA).}
\label{tab:arch-moe}
\footnotesize
\centering
\begin{adjustbox}{center}
\begin{tabular}{@{}lcccccccc@{}}
\toprule
Models
& \makecell{Total \\ Params.}
& \makecell{Activated \\ Params.}
& Layers
& \makecell{Heads \\ (Q / KV)}
& \makecell{Routed / Activated \\ Experts}
& \makecell{Shared \\ Experts}
& \makecell{Context \\ Length} \\
\midrule
A.X K1       & 519B & 33B  & 61 & 64 & 192 / 8 & 1 & 128K \\
\bottomrule
\end{tabular}
\end{adjustbox}
\label{tab:ax}
\end{table}

\subsection{Resource-Constrained Model Design}
A.X K1 was developed over approximately four months as part of the 'Sovereign AI foundation model' project\footnote{The Ministry of Science and ICT(MSIT), Republic of Korea, through the National IT Industry Promotion Agency(NIPA) (Grant No. PJT-25-080042).} funded by the Korean government. Within this constrained timeline, our objective was to deliver a 519B-parameter MoE model that remains competitive with contemporary frontier models. Under a fixed computational budget, we carefully budgeted the available training FLOPs and used these estimates to choose a principled trade-off between model scale and the number of training tokens. These constraints shaped several efficiency-driven design decisions---from the MoE configuration to the training recipe---so that we could maximize capability within a feasible time and compute. While the resulting model inevitably reflects the limits of the available budget, our work demonstrates a practical and reproducible path to building a large-scale MoE model under real-world constraints.

\subsection{Model Configuration}
The A.X K1 model is a large-scale Mixture-of-Experts (MoE) language model with a total of 519B parameters and 33B active parameters, a configuration strategically chosen to maximize knowledge capacity within our hardware constraints while ensuring feasible inference latency. Detailed architectural specifications for A.X K1 are presented in Table~\ref{tab:ax}. For the pre-training compute budget, given a strict training timeline of approximately 75 days on 1,024 H200 GPUs, and assuming a sustained training throughput of $3.84\times10^{14}$ FLOPs/s/GPU (estimated based on efficiency characteristics in~\citep{liu2024deepseek}), the total compute budget was derived as $C \approx 2.548\times10^{24}$ FLOPs. Under this fixed architecture and compute budget, subsequent architectural trade-offs were guided by established scaling principles for MoE models.

Regarding the granularity of expert capacity, we referenced the scaling laws in \citep{tian2025greaterleveragescalinglaws} to balance efficiency with engineering stability. Granularity is defined as $(2\times d_{\text{model}}) / d_{\text{expert}}$, where $d_{\text{model}}$ denotes the FFN hidden size and $d_{\text{expert}}$ the intermediate size of each expert.While a granularity range of approximately 8–12 is reported to be optimal under well-balanced expert routing, this condition is not always achieved in practice. Given $d_{\text{model}} = 7{,}168$, we selected $d_{\text{expert}} = 2{,}048$, resulting in a granularity ($G=7$) slightly below the nominal optimal range. This choice aligns with the observation in \citep{tian2025greaterleveragescalinglaws} that under imperfect expert load balancing, moderately lower granularity values can yield better efficiency and training stability, favoring robustness over idealized balanced assumptions.

The architectural design of A.X K1 adopts several widely used components that have proven effective in large-scale MoE systems~\citep{liu2024deepseek, yang2025qwen3technicalreport, 5team2025glm45agenticreasoningcoding}. These include Multi-head Latent Attention (MLA)~\citep{liu2024deepseek}, which improves Key--Value cache efficiency and reduces memory overhead under long-context settings. Auxiliary-loss-free load balancing is adopted for expert routing, and multi-token prediction (MTP; \citealp{gloeckle2024betterfasterlarge}) is used to support speculative decoding during inference~\citep{leviathan2023fast}. 

\begin{wrapfigure}{r}{0.5\textwidth}
	{
	\includegraphics[width=75mm]{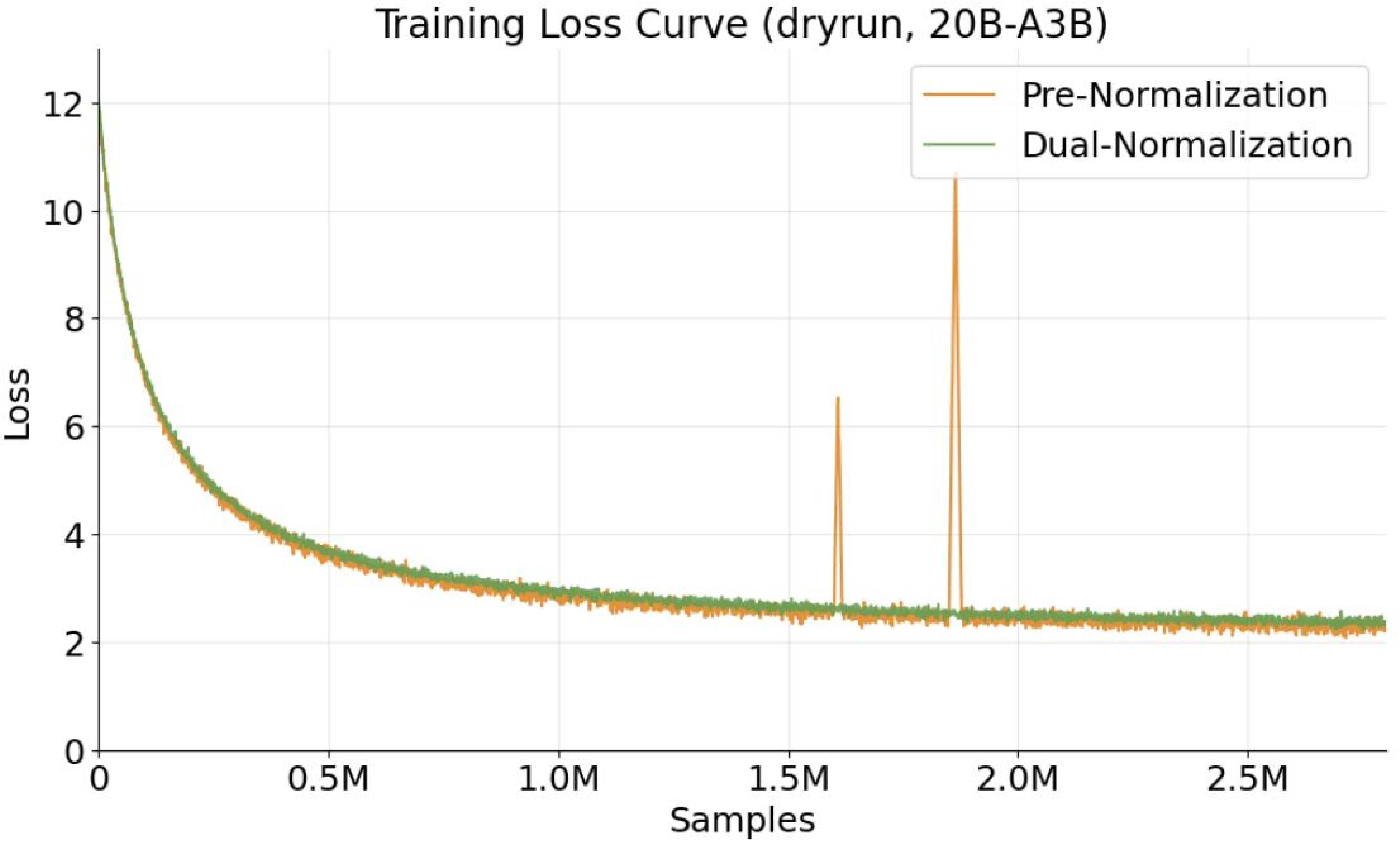}
    \caption{Training loss curves for Pre-Normalization and Dual-Normalization during early training of A.X K1 Light (20B-A3B, small-scale model for dry-run).}
    \label{fig:loss_plot}
	}
\end{wrapfigure}

To improve training stability, we adopt a dual normalization scheme inspired by the design in~\citep{team2025gemma}, applying RMSNorm~\citep{zhang2019rootmeansquarelayer} both before and after the MoE layers. This configuration effectively reduced loss spikes observed during the early stages of training, as shown in Fig.~\ref{fig:loss_plot}. 

Early GLM-4 model series\footnote{https://docs.vllm.ai/en/stable/api/vllm/model\_executor/models/glm4}  closely follow the Gemma architecture, employing post-self-attention normalization, followed by pre-MLP and post-MLP normalizations, resulting in stacked normalization layers. In contrast, we streamlined this design by applying RMSNorm strictly at the input and output of the MoE blocks. This simplified dual-normalization approach maintains training stability while reducing architectural redundancy compared to earlier implementations.

Separately, informed by empirical observations in \citep{kimiteam2025kimik2openagentic}, we configured the model with 64 attention heads and integrated shared dense experts as an operating point that prioritizes inference efficiency by reducing attention overhead while enhancing knowledge sharing. We also evaluated alternating sliding-window attention~\citep{openai2025gptoss}. While small-scale experiments showed promising gains, the method exhibited unreliable behavior with large-scale context parallelism in our pre-training dry runs. Prioritizing training stability for long-context pre-training, we excluded this mechanism from the final architecture.

\subsection{Tokenizer}

We developed a tokenizer designed to efficiently process a diverse training corpus comprising multilingual text, reasoning-oriented data, and code. Employing the standard Byte-level Byte-Pair Encoding (BBPE) algorithm~\citep{sennrich2016neural}, we selected a vocabulary size of 160K to ensure comprehensive coverage across English, Korean, Chinese, Japanese, and Spanish. While recent studies propose Morpheme-aware BBPE~\citep{asgari2025morphbpe}, we opted for the standard BBPE formulation for this project. A detailed analysis of tokenization efficiency is presented in Section~\ref{subsec:tokenizer-eval}.

\paragraph{Design of Vocabulary Size.}
We optimize the vocabulary size based on the \textit{Derivative-based Estimation} from \citep{tao2024scaling}, which suggests a theoretical baseline of approximately 132,500 for our architecture ($N_{nv}=32$B, $d=7168$, indicating non-vocabulary parameters and embedding dimension respectively) under compute-optimal assumptions~\cite{hoffmann2022training}. However, to accommodate our 10T-token training regime, we increased the optimal size from 132,500 to 163,840 (approximately a 25\% increase), based on preliminary empirical observations from smaller-scale models. This consideration motivates our decision to expand the vocabulary size relative to the 102,400-token vocabulary used in our previous model~\citep{SKTAdotX4}. 

This adjustment balances theoretical scaling with practical constraints. First, it accounts for the upward shift in optimal vocabulary size required by data-rich regimes \citep{tao2024scaling} while avoiding the excessive inference latency associated with overly large vocabularies. Second, the selected size is a multiple of 128, ensuring hardware alignment for efficient Tensor Core utilization during embedding and projection operations. This choice represents a robust trade-off between scaling laws and hardware efficiency.

%% file: pretraining.tex
\section{Pre-training}
\label{sec:pretrain}

\subsection{Pre-training Dataset}
\begin{figure}[t!]
    \centering
    \includegraphics[width=\linewidth]{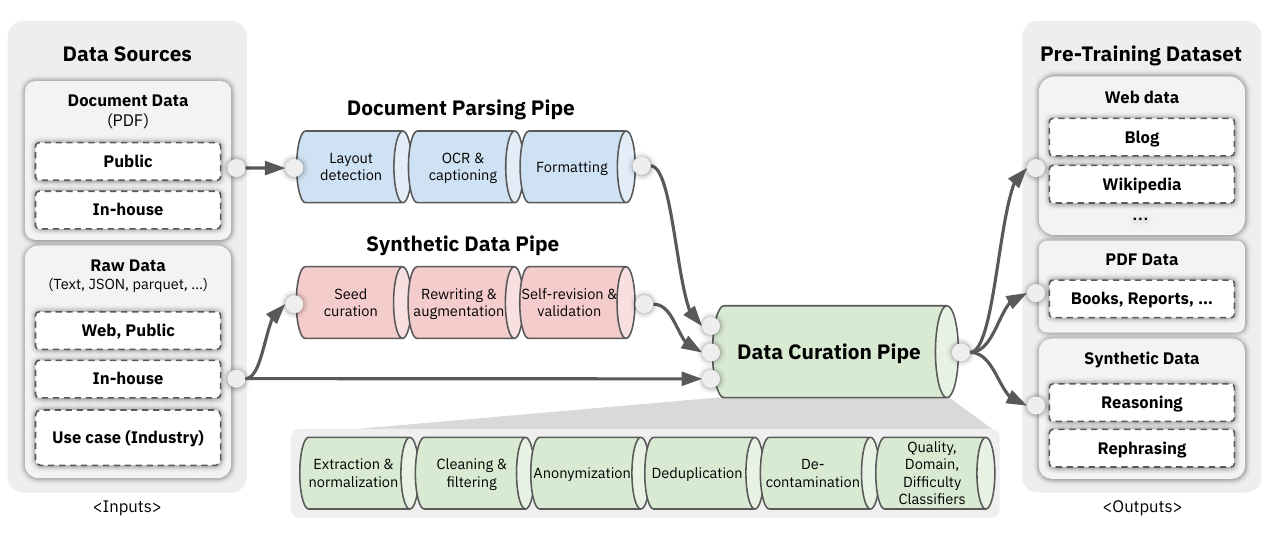}
    \caption{\textbf{A Data Processing Framework for Constructing High-Quality Pre-training Datasets.} The diagram illustrates the transformation of raw data sources (documents and raw data) into a final training corpus through document parsing, synthetic data generation, and a multi-stage curation pipeline (including cleaning, anonymization, deduplication, and diverse classifiers).}
    \label{fig:datapipe-arch}
\end{figure}

We constructed a comprehensive pre-training corpus comprising 10 trillion tokens using a data pipeline shown in Fig.~\ref{fig:datapipe-arch}. This dataset encompasses diverse web corpora alongside high-quality content spanning various domains, including code, STEM (Science, Technology, Engineering, and Mathematics), reasoning, books, and synthetic data. All data sources are refined through the data processing pipeline that utilizes multi-stage quality filtering and fine-grained classifiers for attributes such as domain and difficulty.

To further enrich the corpus, we generated synthetic data using a lightweight language model~\citep{SKTAdotX4Light} by rephrasing scraped Korean web data and designing diverse prompting strategies, including Wikipedia and QA. Additionally, we employed a finetuned VL model~\citep{SKTAdotX4VLLight} to extract text from a massive collection of PDF documents. The extracted text was subsequently refined through a PDF processing pipeline to ensure high quality. Collectively, these augmentation strategies contributed trillions of additional high-quality tokens to the final dataset. We detail these processes in the following section and provide the mixture ratio for each training stage in Appendix~\ref{appendix:pretrain_data}.

\paragraph{Document parsing pipeline.} PDF documents, ranging from academic literature and books to technical reports, contain a wealth of high-quality knowledge but often suffer from structural complexity. To unlock this data, we established a robust document processing pipeline. We use an in-house vision-language model designed to interpret complex visual content specific to Korean. We process this with finetuned layout detection and parsing models, trained in-house to precisely identify and reconstruct diverse document elements (e.g., headers, multi-column text, tables). This pipeline was applied to a massive collection of crawled web documents, internal repositories, and books. By filtering parsing errors and preserving logical structures, we successfully extracted a large-scale, high-quality corpus from sources.

\paragraph{Synthetic data pipeline.} 
We implement a dual synthetic data pipeline designed to enhance both reasoning depth and knowledge breadth. Our generation strategy is categorized into two complementary domains: 
\begin{itemize}
    \item \textbf{Seed corpus-based Synthetic Data.} We focus on transforming existing high-quality text into complex training signals. We first curate a seed corpus comprising \emph{high-difficulty documents from specific sources (e.g., STEM)} and intricate logical texts. Utilizing an ensemble of open-source LLMs, including in-house models~\citep{SKTAdotX4}, we restructure this content into multi-step reasoning chains or agentic workflows. This process effectively converts static information into dynamic, process-oriented data, significantly boosting the model's capabilities in complex problem-solving and agentic planning.
    \item \textbf{Topic-based Synthetic Data.} To systematically elicit the latent \emph{parametric knowledge} stored within teacher models, we employ a topic-driven generation strategy. We aggregate a diverse set of prompts derived from real-world user queries and extract core technical keywords from STEM textbooks. These topics serve as generation seeds, prompting LLMs to synthesize comprehensive explanations and new knowledge contexts that might be sparse in the pre-training corpora.
\end{itemize}

All generated datasets are subsequently processed through the suite of classifiers detailed in our data curation pipeline.

\paragraph{Data curation pipeline.} The quality and distribution of pre-training data are crucial determinants of large language model performance. To ensure our model acquires robust reasoning capabilities and broad world knowledge, we implemented a sophisticated three-stage data curation pipeline: \textbf{Quality Filtering}, \textbf{Domain Classification}, and \textbf{Difficulty Scoring}. We finetune our lightweight Korean text encoder~\citep{SKTAdotXEncoder-base} designed for processing long-context sequences, to train all classifiers.

\begin{itemize}
    \item \textbf{High-Quality Filtering. } To mitigate the impact of noisy and low-information content inherent in web-scale corpora, we employ a multiple filtering process. This process combines heuristic filtering, deduplication, and model-based filtering. This ensures that the base corpus maintains high linguistic coherence, educational value, and information density.
    \item \textbf{Domain-Aware Mixture and Upsampling. } Diversity in training data is essential, but uniform sampling often leads to the under-representation of reasoning-intensive domains. We developed a \emph{Domain Classifier} trained on a taxonomy of academic and general topics. Based on the classifier's output, we construct a weighted data mixture that strategically upsamples reasoning-related domains. In particular, we heavily weigh \emph{STEM}. This re-balancing strategy is designed to enhance the model's logical reasoning and problem-solving skills.
    \item \textbf{Difficulty Scoring for Curriculum Learning. } Beyond domain classification, optimizing the order of data presentation is critical. We introduced a \emph{Difficulty Classifier} that scores documents based on lexical complexity, syntactic structure, and semantic depth. Leveraging these scores, we adopt a curriculum Learning~\citep{bengio2009curriculum}. The training process initiates with simpler, fundamental concepts and progressively introduces higher-difficulty samples. This scheduling facilitates better generalization on complex reasoning tasks.
\end{itemize}

\subsection{Pre-training Process}

To effectively optimize A.X K1 across varying data distributions, we adopt the Warmup-Stable-Decay (WSD) scheduler~\citep{hu2024minicpm} within our multi-stage pre-training pipeline. In our setup, the stages of the WSD scheduler do not directly correspond to the pre-training stages defined by data composition. Instead, the pre-training of A.X K1 is organized into three stages: general knowledge training, advanced knowledge and reasoning training, and long-context extension, based on training objectives and data characteristics. The following sections describe each stage in detail.

\begin{enumerate}
    \item \textit{General Knowledge Stage (Stable Phase):} In the initial pre-training stage, A.X K1 models are trained on a corpus of 7 trillion tokens with a sequence length of 4,096. This phase establishes a robust foundation in both linguistic proficiency and general world knowledge. Aligning with the WSD strategy, we maintain a constant learning rate throughout this stage to maximize knowledge acquisition.
    
    \item \textit{High-Quality Reasoning Stage (Decay Phase):} Following the initial stage, we transition to a specialized phase focused on enhancing reasoning capabilities. This stage utilizes a curated corpus of 1.66 trillion tokens, strictly filtered for high-complexity web data. Furthermore, we incorporate a diverse mixture of domain-specific datasets, including STEM, reasoning-intensive tasks, refined PDF content, and high-quality synthetic data. In this stage, we initiate the learning rate decay to stabilize the model on high-quality distributions.
    
    \item \textit{Long-Context Adaptation:} To enable long-context capabilities, we adopt a multi-stage curriculum that progressively extends the model’s context length from 4K to 16K and subsequently to 32K, using a total of 600 billion tokens. This design is motivated by the demands of agentic tasks, which require long-horizon reasoning, planning, and action over extended contexts. Accordingly, the long-context training stages incorporate data featuring long chains of thought and repository-level code. We maintain a constant learning rate throughout all extension phases.
\end{enumerate}
\subsection{Hyperparameters}

Learning rate and batch size were selected based on the MoE scaling laws proposed by \citep{tian2025greaterleveragescalinglaws}, which were introduced earlier in the context of model size and architecture design. Under these scaling laws, the maximum learning rate is given by $1.1576\times C^{-0.1529}$, which evaluates to approximately $2.147\times 10^{-4}$ for the estimated compute budget; in practice, the peak learning rate was set to $2.1\times 10^{-4}$. The scaling law also recommends a global batch size of $0.0694 \times C^{0.3644}$, corresponding to roughly 54 million tokens.

Since larger batch sizes are generally beneficial for throughput in MoE training, the effective batch size was increased to 64M tokens. During training, the GPU count was later expanded to 1,536, at which point the global batch size no longer aligned with a power-of-two configuration; to maintain high utilization under the updated hardware configuration, the batch size was further adjusted to approximately 75M tokens.

To mitigate gradient spikes and ensure stability during the early training phase, a global batch size ramping-up strategy was applied, linearly increasing the batch size from 2,048 to 16,384 in increments of 1,024 during the warmup period, which spans 72 million samples (approximately 295B tokens with a sequence length of 4,096).

We adopt the AdamW optimizer for pre-training, with momentum parameters set to $\beta_1 = 0.9$ and $\beta_2 = 0.95$. Although alternative optimizers such as Muon~\citep{liu2025muonscalablellmtraining} were evaluated, they were not adopted in this iteration due to the observed degradation in training throughput in our framework. With recent improvements addressing these performance issues, we plan to adopt Muon or its variants in subsequent versions.

In addition to the general optimization hyperparameters, several MoE-specific settings were configured to balance routing stability and expert utilization. The auxiliary loss coefficient for load balancing was set to $1\times10^{-4}$, providing a weak regularization signal without dominating the primary training objective. The top-$k$ scaling factor was set to 2.5 to moderate the effective routing logits and prevent overly sharp expert assignments. The router bias update rate was set to $1\times10^{-3}$ during the earlier stages of training to facilitate adaptation of expert selection; however, this update rate was reduced to zero in Stage~3 to stabilize routing behavior once training shifted toward higher-quality and long-context data.

For MTP, the loss scaling factor was initially set to 0.3 and reduced to 0.1 starting from Stage~2, reflecting a transition from emphasizing auxiliary prediction signals in earlier training to prioritizing the primary language modeling objective in later stages.

\subsection{Parallelism Configuration}

Parallelism configurations were selected on the basis of available hardware capacity and observed communication efficiency. Pre-training was conducted on NVIDIA H200 GPUs with 140\,GB memory per device. Pipeline parallelism (PP) was set to 16, while tensor parallelism was not used, as the available memory capacity was sufficient under this configuration. We employed a distributed optimizer, which partitions optimizer states across data-parallel ranks and achieves memory savings comparable to ZeRO Stage~2~\citep{rajbhandari2020zeromemoryoptimizationstraining}.

Expert parallelism (EP) was fixed at 8. While higher EP degrees are feasible in principle, we did not observe proportional scaling gains beyond this point in our deployment, primarily due to network-communication characteristics rather than algorithmic constraints. In particular, our expert-parallel communication backend (DeepEP) \citep{deepep} benefits from low-level access to high-performance network features (e.g., RDMA-class transports). In managed cloud environments, however, the effective bandwidth and latency can vary depending on the instance type, drivers, and networking configuration (e.g., AWS Elastic Fabric Adapter, EFA) \citep{aws-efa}. Under our configuration, scaling efficiency diminished at higher EP degrees, and thus EP$=8$ provided the best stability--efficiency trade-off. We note that these observations reflect our specific environment and configuration, and may differ under other hardware or networking setups.

Importantly, we incorporated these system-level effects into our end-to-end compute planning. Large-scale profiling revealed that the effective training throughput (FLOPs/s) fell short of our initial estimates under the chosen EP and networking conditions. To preserve the intended total training compute given the project timeline, we increased the training footprint from 1024 to 1536 H200 GPUs while continuing to optimize the communication and parallelization settings. This adjustment reflects a deliberate commitment to deliver a competitive model under practical constraints, backed by measurement-driven planning and efficiency-focused engineering.

Context parallelism was employed to support long-context training as the sequence length increased. Specifically, a context parallel degree of 4 was used when extending the sequence length to 16K, and a degree of 8 was applied for 32K sequences, allowing attention computation to be distributed across devices while keeping per-GPU memory usage within practical limits.

Training was initially launched on 1,024 GPUs and expanded to 1,536 GPUs after several days, with pre-training continuing for approximately 73 days in total. Following the increase in GPU count, the global batch size was scaled from 16,384 to 18,432, corresponding to an increase in tokens per step from approximately 64M to 75M.

Table~\ref{tab:optimization_flops} summarizes the impact of successive system-level optimizations on per-GPU training throughput measured in TFLOPs. The baseline configuration in this table reflects performance observed under the networking constraints of the training environment and should be interpreted as a reference point for the optimization process rather than as a general characterization of expert parallelism scalability. Subsequent improvements were achieved through parallelism reconfiguration, FP8 precision, and fused cross-entropy computation, cumulatively resulting in a substantial increase in effective training throughput.

\begin{table}[t]
\centering
\caption{Per-GPU throughput measured in TFLOPs on NVIDIA H200 GPUs under a sequence length of 4{,}096 tokens and a global batch size of 16{,}384, evaluated on a 64-node cluster in our training environment. The Gain column reports the relative TFLOPs improvement introduced by each optimization compared to the previous configuration.}
\label{tab:optimization_flops}
\begin{tabular}{lrr}
\toprule
\textbf{Optimization} & \textbf{Gain} & \textbf{TFLOPs/GPU} \\
\midrule
Baseline (EP 32, PP 8)            & --        & 95.29  \\
Parallelism reconfiguration (EP 8, PP 16) & +174.2\% & 261.27 \\
FP8 support                    & +10.6\%   & 288.97 \\
Cross-entropy fusion              & +4.4\%   & 301.72 \\
\bottomrule
\end{tabular}
\end{table}

\subsection{Memory Management and Training Efficiency}

Activation recomputation~\citep{korthikanti2022reducingactivationrecomputationlarge} was employed to manage activation memory pressure across all training configurations, including the baseline 4K context length. Recomputation was applied selectively rather than uniformly across the model. Specifically, recomputation was enabled for the up-projection in MLA and for the MLP layers, which together accounted for the dominant portion of activation memory consumption. Other components were left unchanged to avoid unnecessary compute overhead. While full activation recomputation can further reduce peak memory usage, preliminary experiments indicated that it resulted in a substantial slowdown in training throughput, on the order of several tens of percent. Given this trade-off, selective recomputation was chosen as a practical compromise, providing sufficient memory savings while maintaining acceptable training efficiency.

FP8 precision was employed during training, corresponding to the FP8 optimization stage reported in Table~\ref{tab:optimization_flops}. Specifically, the forward pass used the E4M3 format, while the backward pass used E5M2, reflecting the different numerical stability requirements of activations and gradients~\citep{micikevicius2022fp8formatsdeeplearning}. A per-tensor scaling strategy was adopted for FP8 tensors. To preserve numerical stability, the main model parameters and gradients were maintained in FP32, while the exponential moving averages and squared gradients used by the optimizer were stored in BF16. All remaining eligible tensors were represented in FP8. Empirically, the introduction of FP8 resulted in a training throughput improvement on the order of ten percent, consistent with the per-GPU TFLOPs gains shown in Table~\ref{tab:optimization_flops}.

%% file: posttraining.tex
\section{Post-training}
\label{sec:posttrain}
A.X K1 supports user-controlled hybrid inference, where thinking and non-thinking modes are explicitly specified at inference time. Our post-training pipeline is tailored to this setting: it strengthens reasoning when thinking is enabled, while reducing mode confusion to keep non-thinking outputs concise and stable. We also apply a reinforcement learning method that remains stable and compute-efficient for large MoE models, enabling performance improvement without training instability.

\subsection{Post-training Recipe}

Post-training of A.X K1 comprises three stages as follows:

\setlength{\emergencystretch}{2em} 
\begin{enumerate}
    \item \textit{Supervised Fine-tuning:} The initial post-training stage adopts a dual-track strategy comprising Instruct SFT and Reasoning SFT. The former is trained on general instruction datasets (e.g., summarization, rewriting, retrieval-QA, tool use) excluding the \texttt{<think>...</think>} format, while the latter focuses on reasoning-intensive tasks (e.g., math, code) incorporating \texttt{<think>...</think>} reasoning traces. Both models originate from the same pre-trained checkpoint and are fine-tuned independently.

    \item \textit{Think-Fusion SFT:} To design a model capable of selectively employing both thinking and non-thinking modes, we introduce the Think-Fusion SFT framework. This provides a compute-efficient alternative to joint multi-objective fine-tuning. Subsequently, we perform additional fine-tuning on this merged checkpoint. While the merged model inherits strong reasoning priors, it risks mode confusion, potentially over-generating verbose thoughts in non-thinking mode. We mitigate this via a Mode-Overlap Dataset (MOD) strategy: for identical prompts, the training set includes both reasoning-intensive trajectories ($y_{\text{think}}$) and standard direct responses ($y_{\text{non-think}}$). Fine-tuning on this mixed distribution aligns the parameter space, ensuring the model retains deep reasoning capabilities while recovering the versatility to provide concise instructions.
    
    \item \textit{On-policy Reinforcement Learning:} This stage further refines the model’s instruction-following behavior via on-policy reinforcement learning. The training data are drawn exclusively from the instruction-following domain, utilizing the Llama-Nemotron dataset~\citep{bercovich2025llamanemotron}, its translated variants, and internal datasets. Methodologically, we adapt the DAPO framework~\citep{yu2025dapo} by integrating the GSPO loss~\citep{zheng2025gspo}. This modification is specifically designed to enhance on-policy optimization while preserving training stability and computational efficiency.

\end{enumerate}

\subsection{Supervised Fine-tuning}\label{sec:data-sft}

\begin{table}[t]
\centering
\small
\begin{tabular}{ll rr rr}
\toprule
\multirow{2}{*}{Lang.} & \multirow{2}{*}{Domain} &
\multicolumn{2}{c}{Instruct SFT} &
\multicolumn{2}{c}{Reasoning SFT} \\
\cmidrule(lr){3-4}\cmidrule(lr){5-6}
& & Count & Ratio (\%) & Count & Ratio (\%) \\
\midrule
\multirow{7}{*}{En}
& Math                  & 570k   & 19.6 & 555k     & 12.1 \\
& Code                  & 250k   & 8.6  & 625k     & 13.6 \\
& Science/Knowledge     & 376k   & 12.9 & 1{,}049k & 22.8 \\
& Instruction Following & 127k   & 4.4  & 88k      & 1.9  \\
& Tool-use              & 256k   & 8.8  & 329k     & 7.1  \\
& General               & 499k   & 17.2 & 54k      & 1.2  \\
\midrule
\multirow{7}{*}{Ko}
& Math                  & 198k   & 6.8  & 441k     & 9.6  \\
& Code                  & 1k     & 0.0  & 270k     & 5.9  \\
& Science/Knowledge     & 29k    & 1.0  & 1{,}054k & 22.9 \\
& Instruction Following & 104k   & 3.6  & 60k      & 1.3  \\
& Tool-use              & 107k   & 3.7  & 22k      & 0.5  \\
& General               & 389k   & 13.4 & 53k      & 1.2  \\
\midrule
\multicolumn{2}{l}{Total} & 2.91M & 100.0 & 4.60M & 100.0 \\
\bottomrule
\end{tabular}
\caption{Counts and ratios of data used in post-training data mixture.}
\label{tab:data-breakdown}
\end{table}

The supervised fine-tuning (SFT) data used in A.X K1 are divided into two distinct categories based on whether the thinking mode is enabled: reasoning (thinking mode) data and instruction-following (non-thinking mode) data. Reasoning SFT is trained exclusively on thinking mode data, while Instruct SFT relies solely on non-thinking mode instruction data. When the thinking mode is enabled, the response follows the format \texttt{<think>...</think>\{response\}} proposed in ~\citep{deepseekai2025deepseekr1, yang2025qwen3technicalreport}. In contrast, when the non-thinking mode is used, the response follows the format \texttt{</think>\{response\}}. The domain-wise composition and proportions of the datasets used for each SFT stage are summarized in Table~\ref{tab:data-breakdown}. For both stages, we minimize the standard negative log-likelihood loss over the corresponding stage-specific datasets for supervised fine-tuning.

\paragraph{Instruct SFT Data Construction.}
While the English subset primarily utilizes open-source repositories, the Korean component is constructed entirely in-house, employing data curation and augmentation to enhance both scale and quality. To ensure diverse coverage, prompts are aggregated from existing responses and newly curated instruction sets, with concise, direct responses from open-source models. Finally, we uphold rigorous quality standards by filtering candidates through difficulty classification and correctness verification. Formally, we compile these filtered samples into a standard instruction dataset $\mathcal{D}_{\text{non-think}} = \{(x, y_{\text{non-think}})\}$.

\paragraph{Reasoning SFT Data Construction.}
The English thinking mode dataset encompasses a diverse range of open-source resources spanning mathematics, coding, STEM, and tool-use domains, alongside agentic trajectories. Crucially, it incorporates complex coding tasks requiring long-horizon planning (e.g., SWE-style debugging and code repair) as well as long-context scenarios for QA and summarization. To address the scarcity of high-quality Korean reasoning resources, we constructed a comprehensive dataset through a hybrid strategy combining proprietary data curation and strategic augmentation. This corpus integrates manually annotated domain-specific sources, such as mathematics competitions and qualification examinations, with synthetic reasoning trajectories generated via our custom pipelines involving cross-lingual translation and open-source models. This approach enables the creation of a high-density Korean reasoning corpus while preserving task diversity and structural consistency. This process yields the reasoning-intensive dataset $\mathcal{D}_{\text{think}} = \{(x, y_{\text{think}})\}$, where $y_{\text{think}}$ contains explicit reasoning traces.
\subsection{Think-Fusion SFT}
Recent think-hybrid LLMs integrate thinking and non-thinking behaviors into a single checkpoint, but the fusion recipe is highly sensitive: in particular, the mixing ratio between non-reasoning and reasoning supervision varies across models~\citep{yang2025qwen3technicalreport,bae2025exaone4} and can critically affect controllability. Consistent with these observations, prior empirical analysis \citep{wang2025demystifying} shows that mode separation depends on multiple interacting factors (e.g., data ratio and training schedule), often requiring iterative train--evaluate cycles to identify a workable configuration.
To avoid expensive retraining sweeps over model--data mixtures, we propose a Think-Fusion SFT recipe. Specifically, we initialize the model by linearly merging the Instruction SFT model and the Reasoning SFT model using a predefined fusion coefficient.
Although the linearly merged model achieves reasonable performance, the simple model merging approach can cause mode confusion, leading to degraded response quality and format instability (e.g., malformed or unclosed \texttt{<think>} tags). To this end, we fine-tune the merged model on a \textsc{Mode-Overlap} dataset, which pairs identical prompts with both reasoning-intensive and concise response trajectories. 
The alignment step significantly reduces format errors and facilitates flexible switching between the two modes without sacrificing performance.

\paragraph{Linear Model Merging Initialization.} We initially construct a hybrid parameter space by linearly interpolating between a specialized reasoning model ($\theta_{\text{think}}$) and a standard instruction-tuned model ($\theta_{\text{non-think}}$). We define the initialized weights $\theta_{\text{init}}$ as

\begin{equation}
\theta_{\text{init}} = \alpha \cdot \theta_{\text{think}} + (1 - \alpha) \cdot \theta_{\text{non-think}}.
\end{equation}

Empirically, we set the mixing coefficient $\alpha = 0.8$, which was determined in collaboration with \href{https://www.krafton.ai/ko/}{KRAFTON AI}. We found that it strikes a good balance between reducing mode confusion and maintaining generation quality. This high ratio intentionally biases the model towards the reasoning distribution to ensure the retention of complex problem-solving capabilities (i.e., \textit{thinking} features) in the latent space before the subsequent on-policy RL.

\paragraph{Mode-Overlap Dataset Construction.} A critical challenge in training fused models is the discrepancy between the merged weights and the target output distribution. Training solely on reasoning traces ($\mathcal{D}_{\text{think}}$) can force the model to over-generate verbose thought tokens, while training only on standard data ($\mathcal{D}_{\text{non-think}}$) may degrade the merged reasoning capabilities.

To reduce mode confusion, we construct a Mode-Overlap Dataset (MOD), denoted as $\mathcal{D}_{\text{mix}}$. For a given prompt $x$, we curate paired responses covering both modes:

\begin{enumerate}
\item \textbf{Thinking Mode ($y_\text{think}$):} Responses with explicit reasoning chains, drawn from or generated to match the distribution of $\mathcal{D}_{\text{think}}$.
\item \textbf{Non-thinking Mode ($y_\text{non-think}$):} Direct and concise responses aligned with $\mathcal{D}_{\text{non-think}}$.
\end{enumerate}

The final training objective minimizes the negative log-likelihood over the mixed distribution:
%
%
\begin{equation}
\mathcal{L}_{\text{SFT}} = - \sum_{(x, y) \in \mathcal{D}_{\text{mix}}} \log P(y \mid x; \theta_{\text{init}}),
\end{equation}
where $\mathcal{D}_{\text{mix}}$ contains both $(x, y_{\text{think}})$ and $(x, y_{\text{non-think}})$. By supervising the merged model using the overlapping data, we encourage the model to navigate both responses, effectively preventing mode confusion and allowing the model to utilize the strong reasoning priors injected during the merging phase while retaining conversational versatility.

\subsection{On-policy Reinforcement Learning}
The on-policy reinforcement learning (RL) stage is trained on unified instruction-following prompts constructed for paired thinking and non-thinking responses.

Formally, the RL dataset consists of unlabeled prompts, denoted as $\mathcal{D} = \{x_*\}$. To explicitly train both inference modes, we construct two distinct prompt variants for each base instruction $I$: a reasoning-inducing prompt $x_{\text{think}} = [I;$ \texttt{<think>}$]$ and a concise-response prompt $x_{\text{non-think}} = [I;$ \texttt{</think>}$]$. These variants are treated as independent training samples within $\mathcal{D}$. This setup ensures that the optimization process reinforces the desired format and content strictly based on the control tokens appended to the prompt.

For optimization, we adopt GSPO-style sequence-level loss~\citep{zheng2025gspo} within the DAPO on-policy reinforcement learning framework~\citep{yu2025dapo}. As stated in GSPO, unlike standard token-level objectives, which are sensitive to the high variance of individual token likelihoods induced by dynamic expert routing, this sequence-level formulation is particularly robust for MoE architectures.

To further ensure robustness and prevent mode confusion, we incorporate format-aware rewards:
%
%
\begin{equation}
R_{\textrm{total}}(y) = R_{\textrm{correct}}(y) + R_{\textrm{format}}(y),
\end{equation}
where $R_\textrm{correct}(y) \in \{1, -1\}$ reflects the solution accuracy, assigning positive feedback for correctness and negative for errors. Crucially, $R_\textrm{format}(y) \in \{0, -1\}$ functions as a strict penalty term; it assigns $-1$ for incorrect or ambiguous usage of reasoning tokens (e.g., malformed tags) and $0$ for valid structures. These choices ensure reliable user-controlled exposure of reasoning traces without compromising instruction-following quality.

%% file: evaluation.tex
\section{Evaluation}
\label{sec:evaluation}

\subsection{Evaluation Settings}
In evaluating the A.X K1 model, we adopt a benchmark selection strategy that combines (1) representative public benchmarks widely used in recent large language model evaluations and (2) internally constructed benchmarks designed to better reflect Korean-language usage and domain-specific requirements.

The evaluation is organized into seven categories: Math, Code, Knowledge, Korean, Instruction Following (IF), Long-context, and Agent; each targeting a distinct aspect of model capability. Several benchmarks are carefully designed or selected with explicit consideration of potential data contamination and over-optimization, particularly for Korean-language and instruction-following tasks.

By jointly leveraging public and in-house benchmarks, our evaluation aims to provide a balanced and reliable comparison across models, covering both globally established benchmarks and settings that are underrepresented in existing evaluations. Detailed descriptions of each benchmark are provided in Appendix~\ref{subsec:details_of_evaluation_benchmarks}.

\subsection{Evaluation Results of Thinking Mode}

\begin{table}[t!]
\centering
\caption{\textbf{Comparison with large-size frontier LLMs (Thinking Mode).}}
\label{tab:LLM_large_latest_think}
\small
\begin{adjustbox}{width=1.0\textwidth,center}
\begin{tabular}{l l l *{3}{c}}
\toprule
\textbf{Domain} & \textbf{Benchmark} & \textbf{Lang.}
& \makecell{\textbf{A.X K1} \\ (519B-A33B)}
& \makecell{\textbf{DeepSeek-V3.1} \\ (685B-A37B)}
& \makecell{\textbf{GLM-4.6} \\ (357B-A32B)} \\
\midrule

Knowledge
& KMMLU        & Korean & 80.2 & 76.5 & 79.9 \\
& KMMLU-Redux  & Korean & 77.9 & 75.9 & 78.2 \\
& KMMLU-Pro    & Korean & 68.1 & 71.4 & 71.8 \\
& CLIcK        & Korean & 84.9 & 84.5 & 84.9 \\
& KoBALT       & Korean & 48.8 & 59.7 & 59.2 \\
& MMLU-Pro     & English & 81.5 & 85.1 & 82.9 \\
& GPQA Diamond & English & 74.0 & 77.9 & 78.0 \\

\midrule
Instruction Following
& IFBench {\tiny(prompt-loose)}                        & English & 64.7 & 41.5 & 43.4 \\
& IFEval {\tiny(prompt-strict)}   & English & 80.4 & 84.4 & 86.1 \\
& IFEval-ko {\tiny(prompt-strict)}& Korean  & 81.0 & 79.2 & 85.8 \\

\midrule
Math
& AIME25    & English & 89.8 & 88.4 & 86.0 \\
& AIME25-ko & Korean  & 86.9 & 78.1 & 75.9 \\
& HRM8K     & Korean  & 89.4 & 84.3 & 83.9 \\
& KMO25     & Korean  & 93.0 & 86.4 & 96.6 \\

\midrule
Code
& LiveCodeBench v6 {\tiny(25.2.1 - 25.5.1)} & English & 75.8 & 69.5 & 76.0 \\
& LiveCodeBench-ko & Korean & 73.1 & 66.2 & 55.9 \\
& HumanEval+       & English & 87.2 & 86.0 & 83.5 \\
& HumanEval+ ko    & Korean  & 90.2 & 93.9 & 86.0 \\
& MBPP+            & English & 93.0 & 99.2 & 98.9 \\
& SciCode          & English & 32.4 & 39.1 & 38.4 \\

\midrule
Long Context
& AA-LCR               & English & 36.0 & 53.3 & 54.3 \\
& Humanity's Last Exam & English & 8.6  & 13.0 & 13.3 \\

\midrule
Agent
& \(\tau^2\) Telecom & English & 58.1 & 37.4 & 70.5 \\

\bottomrule
\end{tabular}
\end{adjustbox}
\end{table}

To validate the effectiveness of A.X K1, we conducted a comprehensive evaluation across diverse domains, including Knowledge, Instruction Following, Math, Code, Long Context, and Agentic capabilities. We compared our model against leading open-weight models, DeepSeek-V3.1 and GLM-4.6, as shown in Table~\ref{tab:LLM_large_latest_think}.

It is important to note that A.X K1 was developed under a significantly more constrained computational budget compared to the baseline models. Despite this disparity in training resources, A.X K1 demonstrates competitive performance in several key benchmarks. Notably, in the Math domain, our model achieved a score of 89.8 on AIME25, outperforming both DeepSeek-V3.1 (88.4) and GLM-4.6 (86.0). Furthermore, in Korean-specific knowledge tasks such as KMMLU, A.X K1 recorded 80.2, proving its superior understanding of the target language context.

While the overall performance across broader benchmarks, particularly in Long Context and complex Coding tasks, shows a slight gap compared to the SOTA models, this can be primarily attributed to the difference in pre-training scale and compute resources. However, the fact that A.X K1 secures State-of-the-Art results in specific high-reasoning tasks (Math) and language-specific benchmarks (Korean) suggests that our architecture possesses strong potential for scaling.

\begin{table}[t!]
\centering
\caption{\textbf{Comparison with large-sized frontier LLMs (Non-Thinking Mode).}}
\label{tab:LLM_large_latest_nonthink}
\small
\begin{adjustbox}{width=0.8\textwidth,center}
\begin{tabular}{l l c c c}
\toprule
\textbf{Domain} & \textbf{Benchmark}
& \makecell{\textbf{A.X K1} \\ (519B-A33B)}
& \makecell{\textbf{DeepSeek-V3.1} \\ (685B-A37B)}
& \makecell{\textbf{GLM-4.6} \\ (357B-A32B)} \\
\midrule

Knowledge
& KMMLU          & 73.0 & 78.7 & 77.7 \\
& KMMLU-Redux    & 68.3 & 75.9 & 73.4 \\
& KMMLU-Pro      & 60.3 & 67.9 & 68.2 \\
& CLIcK          & 77.2 & 80.9 & 77.9 \\

\midrule

Instruction Following
& IFBench        & 44.3  & 37.8 & 36.7 \\
& IFEval        & 78.6  & 82.7 & 87.2 \\

\midrule

Code
& HumanEval+     & 79.9 & 87.8 & 89.0 \\
& HumanEval+ ko  & 75.6 & 86.6 & 92.1 \\
& MBPP+          & 85.7 & 92.6 & 94.2 \\

\bottomrule
\end{tabular}
\end{adjustbox}
\end{table}

\subsection{Evaluation Results of Non-Thinking Mode}

In non-thinking mode, A.X K1 shows mixed performance across instruction-following, knowledge, and code benchmarks shown in Table~\ref{tab:LLM_large_latest_nonthink}. It achieves the highest score on IFBench among the compared models, while scoring lower than DeepSeek-V3.1 and GLM-4.6 on IFEval. Across knowledge benchmarks (KMMLU, KMMLU-Redux, KMMLU-Pro), A.X K1 consistently records lower scores than both comparison models, and remains slightly below DeepSeek-V3.1 on CLIcK while being comparable to GLM-4.6. In code generation benchmarks (HumanEval+, HumanEval+ ko, MBPP+), A.X K1 underperforms both DeepSeek-V3.1 and GLM-4.6, reflecting the impact of training and inference compute budgets under non-thinking inference.

\subsection{Quantitative Analysis of Tokenization Efficiency}
\label{subsec:tokenizer-eval}
We report quantitative evaluations of the proposed tokenizer. Tokenization efficiency was measured across multiple languages and domains, including natural language, reasoning, mathematics, and code, using average token count as the primary metric. The tokenizer consistently reduced token counts across languages, achieving particularly strong efficiency in Korean. Improvements were also observed in English and code, demonstrating efficiency competitive with English SOTA models. Table~\ref{tab:tokenizer-efficiency} reports the average token counts across languages and domains.

\begin{table}[t]
\centering
\caption{Tokenization Efficiency Comparison (Average Tokens per Sample). \textbf{Bold} denotes the best efficiency, and \underline{Underline} indicates the second-best ones.}
\label{tab:tokenizer-efficiency}
\begin{tabular}{lrrr}
\toprule
\textbf{Dataset} &
\textbf{A.X K1} &
\textbf{OpenAI o200k} &
\textbf{Qwen3}  \\
\midrule
Korean (General)       & \textbf{526.72} & \underline{811.71}   & 905.82    \\
Korean (Reasoning)     & \textbf{5520.06} & \underline{7273.03}  & 7890.16   \\
Korean (Math)          & \textbf{812.99} & \underline{1173.98}  & 1297.56   \\
English (General)      & \underline{792.28} & \textbf{769.31} & 794.79    \\
English (Reasoning)    & \underline{6052.71} & \textbf{5872.95} & 6192.77   \\
Chinese                & \underline{1050.24} & 1069.66  & \textbf{914.05}  \\
Japanese               & \textbf{871.30} & 1040.50  & \underline{935.87}    \\
Spanish                & \underline{1059.73} & \textbf{937.10} & 1102.41  \\
Code                   & \underline{1693.33} & \textbf{1680.74} & 1694.26   \\
\bottomrule
\end{tabular}
\end{table}

%% file: conclusion.tex
\section{Limitations}
While A.X K1 demonstrates competitive performance and validates our efficient training and post-training recipe, several limitations remain, largely driven by real-world constraints and stability-focused engineering trade-offs. These limitations also delineate a clear roadmap for future iterations.

\begin{itemize}
    \item \textbf{Infrastructure Constraints on Scaling:} Our training environment imposed system-level constraints on parallelism. In particular, we capped Expert Parallelism (EP) at 8 because scaling efficiency diminished beyond this point under our networking configuration, reflecting communication characteristics rather than algorithmic limits. This restricted our ability to explore finer-grained expert partitioning and higher EP degrees that may yield additional capacity. Future work will benefit from improved interconnect characteristics and tighter integration between expert-parallel communication and the underlying network stack.

    \item \textbf{Optimization under Resource Constraints:} To maximize attainable performance under fixed time and GPU budget constraints, we selected model configurations guided by theoretical scaling laws. Despite parameter-level optimization, these constraints led to modest performance shortfalls relative to similarly sized models trained with larger compute budgets. We expect that increasing available compute will allow us to relax these constraints and further improve model performance.

    \item \textbf{Scope of Modalities:} A.X K1 is currently text-only. While it exhibits strong reasoning performance, it lacks native multimodal understanding. We plan to extend the model with native multimodal capabilities in future works.
\end{itemize}

\section{Conclusion}
In this work, we presented A.X K1, a 519B-parameter MoE language model that bridges high-capacity reasoning with practical inference efficiency. Under fixed and time-bounded resources, we followed vocabulary and MoE scaling principles and used FLOP-based budgeting to select a principled trade-off between model scale and training tokens, demonstrating that careful system-aware design can yield globally competitive capability.
We also introduced \textit{Think-Fusion}, a post-training recipe that combines dual-track SFT with GSPO-style on-policy reinforcement learning to enable explicit, user-controllable switching between \textit{thinking} and \textit{standard} modes within a single unified checkpoint. Across mathematics, coding, and general-knowledge benchmarks in both English and Korean, our evaluations show that A.X K1 is competitive with strong open-source baselines.

Beyond the theoretically optimal architecture defined by scaling laws, this project focuses on the system-level engineering required to translate that design into effective large-scale MoE training under real-world infrastructure constraints. Empirical profiling revealed systematic gaps between theoretical and realized throughput, driven by communication overhead, load imbalance, and execution-level inefficiencies. To preserve the intended training compute and schedule, we expanded the cluster from 1,024 to 1,536 H200 GPUs and iteratively optimized stability, utilization, and throughput. This approach enabled stable convergence and predictable training behavior within fixed resource and timeline constraints.

Looking ahead, we will prioritize native multimodal capabilities and continued scaling toward the trillion-parameter regime. Building on the system and optimization insights established here, we aim to further advance both model capability and the efficiency of large-scale training and inference.

%% file: authors.tex
\section{Contributors}
All authors are listed alphabetically by last name, then first name. Names marked with an asterisk (*) indicate authors whose affiliations differ from their affiliation at the time of this work.

\subsection*{Core Contributors}
Sung Jun Cheon, Jaekyung Cho*, Seongho Choi, Hyunjun Eun, Seokhwan Jo, Jaehyun Jun*, Minsoo Kang, Jin Kim, Jiwon Kim, Minsang Kim, Seungsik Kim, Sungwan Kim, Tae Yoon Kim, Youngrang Kim, Hyeongmun Lee, Sangyeol Lee, Sungeun Lee, Youngsoon Lee, Yujin Lee, Seongmin Ok, Chanyong Park*, Hyewoong Park*, Junyoung Park, Hyunho Yang, Subin Yi

\subsection*{Contributors}
Dhammiko Arya, Soohyun Bae, Dongyeon Cho, Seungmo Cho, Sangho Choi, Yongseok Choi, Gyoungeun Han, Yong-jin Han, Seokyoung Hong, Hyeon Hwang*, Wonbeom Jang, Minjeong Ju, Wonjin Jung, Keummin Ka*, Sungil Kang, Dongnam Kim, Jonghwi Kim*, Joonghoon Kim, SaeRom Kim, Sangjin Kim, Seongwon Kim, Youngjin Kim, Seojin Lee, Sunwoo Lee, Taehoon Lee, Chanwoo Park*, Sohee Park, Sooyeon Park, Yohan Ra, Sereimony Sek, Seungyeon Seo, Gun Song, Sanghoon Woo*, Janghan Yoon*, Sungbin Yoon

\subsection*{Acknowledgement}
This work was supported by the Ministry of Science and ICT(MSIT), Republic of Korea, through the National IT Industry Promotion Agency(NIPA) (Grant No. PJT-25-080042). This research was conducted as part of the Sovereign AI Foundation Model Project(Data Track), organized by the Ministry of Science and ICT(MSIT) and supported by the National Information Society Agency(NIA), S.Korea. (Grant No. 2025-AIData-WII43). We acknowledge \href{https://www.krafton.ai/ko/}{KRAFTON AI}---in particular, Gyeongman Kim, Junhyuck Kim, Haechan Kim, and Beongjun Choi---for their collaboration in the development of the Think-Fusion SFT recipe.

\clearpage

%% file: appendix.tex
\section{Appendix}

\subsection{Pre-train Data Category Mixture}\label{appendix:pretrain_data}

\begin{table}[h]
\centering
\caption{Dataset Composition by Category Across Training Stages.}
\label{tab:dataset-composition-stages}
\begin{tabular}{l r r r}
\toprule
\textbf{Category} & \textbf{Stage 1 (\%)} & \textbf{Stage 2 (\%)} & \textbf{Stage 3 (\%)} \\
\midrule
Web                   & \textbf{53.41} & \textbf{39.79} & \textbf{33.32} \\
Code                  & \textbf{17.13} & \textbf{29.42} & \textbf{25.50} \\
Encyclopedia          & \textbf{14.84} & 0.64  & 0.72  \\
Q\&A                  & \textbf{13.04}  & \textbf{8.95}  & 5.38  \\
Books                 & 0.77  & 4.35  & 3.70  \\
Academic Literature   & 0.66  & \textbf{7.23}  & 3.87  \\
Mathematics           & --    & \textbf{5.41} & \textbf{9.24} \\
STEM                  & --    & 2.47  & \textbf{9.07} \\
Reasoning             & --    & 0.82  & \textbf{8.32} \\
Others                & 0.16  & 0.91  & 0.87 \\
\midrule
Total                 & 100.00 & 100.00 & 100.00 \\
\bottomrule
\end{tabular}
\end{table}

\subsection{Details of Evaluation Benchmarks}
\label{subsec:details_of_evaluation_benchmarks}

\paragraph{Math}
For English benchmarks, we use AIME25~\citep{aime25}, a prestigious high-school–level mathematics competition problem set drawn from the 2025 American Invitational Mathematics Examination (AIME).  Korean math performance is evaluated using AIME25-ko, a Korean translation of AIME25, and HRM8K~\citep{ko2025understand}. All math benchmarks are evaluated based on the correctness of the final numerical answer, using a chat-based chain-of-thought (Chat CoT) prompting setup that encourages sufficient intermediate reasoning. We evaluate AIME25 using a maximum context length of 128K tokens. The evaluation prompt and usage examples for AIME25 and HRM8K are shown in Fig.~\ref{fig:aime25_chat} and Fig.~\ref{fig:hrm8k_chat}, respectively.

\begin{tcolorbox}[colback=gray!10,colframe=black,title=Evaluation Prompt for AIME25,fontupper=\small]
Solve the following math problem efficiently and clearly.  The last line of your response should be of the following format: 'Therefore, the final answer is: \$\textbackslash boxed\{ANSWER\}\$. I hope it is correct' (without quotes) where ANSWER is just the final number or expression that solves the problem. Think step by step before answering.\par
\vspace{\baselineskip}
\textcolor{gray}{Find the sum of all integer bases $b>9$ for which $17_b$ is a divisor of $97_b.$}
\end{tcolorbox}
\noindent\begin{minipage}{\textwidth}
\captionof{figure}{One example of evaluation prompt for AIME25. The model’s response is evaluated by comparing the \texttt{ANSWER} field against the ground-truth answer.}\label{fig:aime25_chat}
\end{minipage}

\begin{tcolorbox}[colback=gray!10,colframe=black,title=Evaluation Prompt for HRM8K,fontupper=\small]
다음 수학 문제를 효율적이고 명확하게 풀어주세요. 마지막 줄의 답변은 다음 형식이어야 합니다: `따라서 최종 답은: \$\textbackslash boxed\{ANSWER\}\$. 맞기를 바랍니다.' (따옴표 없이, ANSWER는 문제를 해결한 최종 숫자나 표현식입니다). 문제를 풀기 전에 한 단계씩 생각해주세요.\par
\vspace{\baselineskip}
\textcolor{gray}{Janet의 오리는 하루에 16개의 알을 낳습니다. 그녀는 매일 아침으로 3개를 먹고, 친구들을 위해 머핀을 구울 때 4개를 사용합니다. 남은 계란은 매일 농산물 시장에서 신선한 오리 알 하나당 2달러에 판매합니다. 그녀는 매일 농산물 시장에서 얼마를 버나요?}
\end{tcolorbox}
\noindent\begin{minipage}{\textwidth}
\captionof{figure}{One example of evaluation prompt for HRM8K. The model’s response is evaluated by comparing the \texttt{ANSWER} field against the ground-truth answer.}\label{fig:hrm8k_chat}
\end{minipage}

\paragraph{Code}
We use HumanEval+, HumanEval+ko, MBPP+, LiveCodeBench~\citep{jain2024livecodebench}, and LiveCodeBench-ko. Generated code is executed using a Python interpreter, and correctness is determined by matching execution results with expected outputs. The LiveCodeBench series includes recently updated problems, reducing the risk of data contamination; we evaluate only the February, March, and April 2025 subsets. All LiveCodeBench evaluations are conducted with a context length of 128K tokens.

\paragraph{Knowledge}
For English knowledge benchmarks, we use MMLU~\citep{hendryckstest2021}, MMLU-Pro~\citep{wang2024mmlupro}, GPQA Diamond~\citep{rein2024gpqa}, and Humanity's Last Exam~\citep{phan2025humanity}. MMLU assesses undergraduate-level knowledge across diverse academic fields, while MMLU-Pro increases difficulty by expanding the number of answer choices. GPQA Diamond consists of high-difficulty, graduate-level questions in physics, chemistry, and biology, and HLE evaluates advanced knowledge and reasoning capabilities. MMLU and MMLU-Pro are evaluated using the chat-based chain-of-thought (Chat CoT) prompting. The evaluation prompts and usage examples for MMLU and MMLU-Pro are shown in Figures~\ref{fig:mmlu_chat} and \ref{fig:mmlu_pro_chat}, respectively.

\begin{tcolorbox}[colback=gray!10,colframe=black,title=Evaluation Prompt for MMLU,fontupper=\small]
Given the following question and four candidate answers (A, B, C and D), choose the best answer.\par
\vspace{\baselineskip}
\textcolor{gray}{Question: What is the greatest common divisor of ?}\par
\textcolor{gray}{A. 1}\par
\textcolor{gray}{B. 2049}\par
\textcolor{gray}{C. 2048}\par
\textcolor{gray}{D. 2047}\par
\vspace{\baselineskip}
- For simple problems:\par
Directly provide the answer with minimal explanation.\par
\vspace{\baselineskip}
- For complex problems:\par
Use this step-by-step format:\par
\#\# Step 1: [Concise description]\par
[Brief explanation]\par
\#\# Step 2: [Concise description]\par
[Brief explanation]\par
\vspace{\baselineskip}
Regardless of the approach, always conclude with:\par
The best answer is [the\_answer\_letter].\par
where the [the\_answer\_letter] is one of A, B, C or D.\par
\vspace{\baselineskip}
Let's think step by step.
\end{tcolorbox}
\noindent\begin{minipage}{\textwidth}
\captionof{figure}{One example of an evaluation prompt for MMLU. The model's response is evaluated by comparing the \texttt{[the\_answer\_letter]} field against the ground-truth answer.}\label{fig:mmlu_chat}
\end{minipage}

\begin{tcolorbox}[colback=gray!10,colframe=black,title=Evaluation Prompt for MMLU-Pro,fontupper=\small]

Answer the following multiple-choice question. The last line of your response should be in the following format: 'Answer: A/B/C/D' (e.g. 'Answer: A').\par
\vspace{\baselineskip}
\textcolor{gray}{Functions of the law include all but which of the following?}\par
\vspace{\baselineskip}
\textcolor{gray}{A) defining the limits of government power}\par
\textcolor{gray}{B) regulating the use of public spaces}\par
\textcolor{gray}{C) keeping the peace}\par
\textcolor{gray}{D) maximizing individual freedom}\par
\textcolor{gray}{E) maintaining order and stability}\par
\textcolor{gray}{F) preventing environmental degradation}\par
\textcolor{gray}{G) providing a basis for compromise}\par
\textcolor{gray}{H) promoting social justice}\par
\textcolor{gray}{I) promoting the principles of the free enterprise system}\par
\textcolor{gray}{J) encouraging economic growth}\par
\end{tcolorbox}
\noindent\begin{minipage}{\textwidth}
\captionof{figure}{An example MMLU-Pro evaluation prompt. Model outputs are scored by parsing the final \texttt{Answer:} field and matching the selected option with the ground-truth answer.}\label{fig:mmlu_pro_chat}
\end{minipage}

\paragraph{Korean}
For Korean knowledge evaluation, we use KMMLU~\citep{son2024kmmlu}, KMMLU-Redux~\citep{hong2025kmmluredux}, KMMLU-Pro~\citep{hong2025kmmluredux}, CLIcK~\citep{kim2024click}, and KoBALT~\citep{shin2025kobalt}. KMMLU is the Korean counterpart of MMLU, while KMMLU-Redux corrects and refines errors in the original KMMLU. KMMLU-Pro includes questions based on the Korean national professional qualification exams. CLIcK and KoBALT evaluate Korean knowledge and reasoning. For KMMLU, we adopt the chat-based chain-of-thought (Chat CoT) evaluation protocol proposed in \citep{dubey2024llama}. In contrast, CLIcK is evaluated in a standard zero-shot setting, consistent with the evaluation protocol used for pre-trained models.

\paragraph{Instruction Following}
This category evaluates the model’s ability to follow given instructions accurately. We use IFEval~\citep{ifeval2023}, IFEval-ko, and IFBench~\citep{pyatkin2025generalizing}, which assess compliance with explicit constraints such as output format, length limits, required expressions, or language usage. IFEval is evaluated using predefined Python scripts, while IFEval-ko is an in-house Korean translation of IFEval. All benchmarks are evaluated using rule-based criteria with predefined evaluation protocols.

\paragraph{Long-context}
Long-context understanding is evaluated using the AA-LCR~\citep{artificialanalysis2025lcr} benchmark. This benchmark evaluates the model’s ability to maintain coherence and retain key information when generating responses based on long input contexts. All evaluations are conducted with a maximum context length of 128K tokens, reflecting the model’s long-context capability.

\paragraph{Agent}
$\tau^2$-Bench Telecom~\citep{barres2025tau2} evaluates agent performance in the Telecom domain, such as customer service interactions.